\documentclass{Interspeech2024}
\usepackage{amsmath}
\usepackage{graphicx}
\usepackage{float}
\usepackage{algorithm}
\usepackage{algpseudocode}
\usepackage{multirow}

\interspeechcameraready

\title{Neuromorphic Keyword Spotting with\\Pulse Density Modulation MEMS Microphones}

\name[affiliation={1}]{Sidi Yaya Arnaud}{Yarga}
\name[affiliation={1}]{Sean U N}{Wood}

\address{
  $^1$Department of Electrical and Computer Engineering, Université de Sherbrooke, Sherbrooke, QC, Canada
  }
\email{sidi.yaya.arnaud.yarga@usherbrooke.ca, sean.wood@usherbrooke.ca}


\begin{document}

\maketitle

\begin{abstract}
     The Keyword Spotting (KWS) task involves continuous audio stream monitoring to detect predefined words, requiring low energy devices for continuous processing. Neuromorphic devices effectively address this energy challenge. However, the general neuromorphic KWS pipeline, from microphone to Spiking Neural Network (SNN), entails multiple processing stages. Leveraging the popularity of Pulse Density Modulation (PDM) microphones in modern devices and their similarity to spiking neurons, we propose a direct microphone-to-SNN connection. This approach eliminates intermediate stages, notably reducing computational costs. The system achieved an accuracy of 91.54\% on the Google Speech Command (GSC) dataset, surpassing the state-of-the-art for the Spiking Speech Command (SSC) dataset which is a bio-inspired encoded GSC. Furthermore, the observed sparsity in network activity and connectivity indicates potential for remarkably low energy consumption in a neuromorphic device implementation.
\end{abstract}

\section{Introduction}
Keyword Spotting (KWS) is the process of identifying predefined words or expressions within a stream of spoken language. Achieving this requires continuous monitoring, necessitating devices with low energy consumption \cite{chen2014small}. To address this demand, various avenues have been explored in the literature to enhance the energy efficiency of KWS devices \cite{lopez2021deep}. Notably, neuromorphic devices have emerged as more effective than traditional approaches, leveraging their bio-inspired binary and event-driven nature \cite{blouw2019benchmarking,yilmaz2020deep}.

Neuromorphic KWS systems typically involve multiple processing steps, including analog-to-digital conversion (ADC), feature extraction, spike encoding, and the utilization of Spiking Neural Networks (SNN). In certain studies, the audio captured by microphones undergoes processing through spiking cochlear models  \cite{cramer2020heidelberg, jimenez2016binaural}. These models emulate the functioning of the biological auditory system, extracting time-frequency features and transforming them into spike format.

Micro-Electromechanical Systems (MEMS) microphones are widely used in various devices using KWS like smartphones, tablets, laptops and automotive \cite{shah2019design}.
Certain modern MEMS microphones adopt Pulse Density Modulation (PDM) via a sigma-delta converter to encode analog signals directly at the sensor, generating a binary stream \cite{maluf2004introduction, kite2012understanding}. These PDM signals bear resemblance to densely timed spikes or bursts in neural communications \cite{sevuktekin2019signal}. Building upon this association, previous works have introduced spike-based systems to adapt PDM signals for direct use by Neuromorphic Auditory Sensors (NAS) \cite{gutierrez2023interfacing} and proposed sound classification systems utilizing Spike Continuous Time Neurons (SCTN) for PDM signal preprocessing \cite{bensimon2021using}. However, these approaches still add supplementary processing stages between the microphone and the SNN.

In this paper, we present a novel approach\footnote{A PyTorch implementation of the proposed approach is made available online at: https://github.com/NECOTIS/Keyword-Spotting-with-PDM} to Keyword Spotting by utilizing the PDM-encoded signals from MEMS microphones directly as input to an SNN. This strategy bypasses traditional preprocessing steps, resulting in a notable reduction in computational requirements. Additionally, we contribute to the understanding of the similarity between PDM signals and Integrate and Fire (I\&F) neurons. 
Furthermore, to validate the effectiveness of our approach with the Google Speech Command (GSC) dataset \cite{warden2018speech}, we introduce a parallel Pulse-Code Modulation (PCM) to PDM conversion algorithm, allowing fast conversion and thereby accelerating the training process.

\section{Proposed Method}
\begin{figure*}[!t]
\centering
\includegraphics[width=0.8\linewidth]{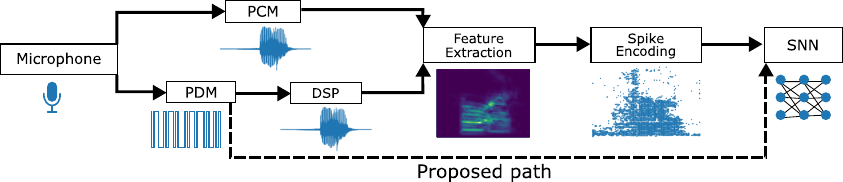}
\caption{General neuromorphic Keyword Spotting pipeline and proposed shortcut.}
\label{img:method}
\end{figure*}

In the context of achieving a neuromorphic KWS system, the typical process involves digitizing the analog audio signal captured by the microphone. Subsequently, this digitized signal undergoes a feature extraction procedure. The extracted features are then encoded into spikes and fed into the classifier, often represented by an SNN as illustrated in Figure \ref{img:method}.

\subsection{Analog-to-Digital Conversion}
The expected audio format is usually a Pulse-Code Modulated (PCM) signal. PCM is a method used to digitally represent analog signals by sampling the amplitude at uniform intervals, and quantizing each sample to the nearest value within a range of digital steps \cite{kite2012understanding}. In PCM, signal fidelity hinges on the sampling rate (e.g., 48 kHz) and the quantization bit depth (e.g., 16 bits). Although certain MEMS microphones produce PCM audio via the Inter-IC Sound interface, PDM stands out as the prevailing digital MEMS microphone interface \cite{lewis2013analog}.

Pulse Density Modulation generates a single-bit stream with a high sampling rate. Within a PDM signal, the density of pulses reflects the amplitude of the analog signal \cite{kite2012understanding}. The fidelity of such a signal depends on the oversampling ratio. For instance, if a desired sampling rate of 48 kHz is targeted, with a 64 oversampling ratio the PDM signal will be sampled at 3 MHz. To convert a PDM signal into PCM audio, a decimation process can be executed within a Digital Signal Processor (DSP) via lowpass filtering, downsampling, and quantization.

\subsection{Feature Extraction}
The feature extraction stage commonly entails a Time-Frequency decomposition, often accomplished through methods like Short Time Fourier Transform (STFT) or perceptually-motivated approaches such as Mel-frequency cepstral coefficients (MFCC). A comprehensive exploration of feature extraction methodologies is provided in \cite{lopez2021deep}. Subsequently, these features are encoded into spikes.
In the literature, bio-inspired cochlear models have been proposed to extract features, emulating the processes observed in the biological inner ear. For instance, in \cite{cramer2020heidelberg}, the audio signal undergoes processing via a hydrodynamic basilar membrane model, facilitating time-frequency decomposition. This is followed by the utilization of hair cell and bushy cell models to enable spike conversion.

\subsection{Spiking Neural Network}
In the final stage, the SNN receives the spike-encoded, extracted features as input and produces a decision regarding the keyword detection. The literature extensively explores various spiking neuron types and network architectures, including Recurrent Neural Networks and Convolutional Neural Networks \cite{dampfhoffer2022investigating, bittar2022surrogate, sadovsky2023speech}.

\subsection{Proposed Approach}
In this paper, we present a novel shortcut, involving the direct input of the PDM signal from the microphone into the SNN as illustrated in Figure \ref{img:method}. The inherently binary nature of the PDM signal aligns well with SNN processing. This approach provides the notable advantage of bypassing all intermediate steps, thereby reducing the computational costs of the overall system. To this end, we address various challenges related to processing signals with high sampling rates and learning to extract relevant features.

The following section outlines the proposed experimental setup, encompassing the SNN architecture and the dataset employed.

\section{Experimental Setup}
\subsection{Network}
\label{sec:network}
To address the challenges posed by the temporal nature and high sampling rate of the PDM-encoded input signal, we opted for the utilization of 1D convolutional SNN. This choice is grounded in their demonstrated capacity to effectively learn representations from temporal audio signals \cite{abdoli2019end}. Within the network, we employed the Delta variant of the Parallelizable Leaky Integrate-and-Fire neuron (ParaLIF-D) \cite{yarga2024accelerating}. ParaLIF-D, a fast and parallel spiking neuron derived from the Leaky Integrate-and-Fire (LIF) neuron, demonstrated remarkable efficiency in processing extended input sequences rapidly through parallelization over time, thereby contributing to accelerated training.
The network topology consists of 5 layers, described as follows :
\begin{itemize}
    \item \textbf{Layer 1}: 128 ParaLIF-D neurons, 1D convolution, kernel size of \(3*\alpha\) , stride of \(\lfloor3*\alpha/2 \rfloor\), where \(\alpha\) is the oversampling ratio.
    \item \textbf{Layers 2, 3, 4}: 128 ParaLIF-D neurons, 1D dilated convolution, kernel size of 3, stride of 3, dilation factor of 2.
    \item \textbf{Layer 5}: 35 Leaky Integrator (LI) neurons that do not produce spikes. The cumulative sum of their membrane potential over time is employed for predictions, following the methodology presented in \cite{bittar2022surrogate}.
\end{itemize}

Furthermore, all hidden layers are followed by a random fixed axonal delay, akin to the approach outlined in \cite{sadovsky2023speech}. The delay for each neuron is uniformly chosen from the range [0-30] time steps, with the time step value depending on the current layer's input sampling rate. Additionally, layers 3 and 4 incorporate a recurrent version of ParaLIF, as detailed in \cite{yarga2024accelerating}, with the modification that the recurrent term is obtained by passing the hidden membrane potential through a Rectified Linear Unit (ReLU) function.

Finally, the surrogate gradient \cite{neftci2019surrogate} is employed for error backpropagation. The optimization process utilizes the Adamax optimizer, initialized with a learning rate of 0.002. A ``reduce on plateau" scheduler is implemented, with a reduction factor of 0.7 and a patience value set at 10.

\subsection{Dataset}

For our experiments, we utilized the GSC dataset v2 \cite{warden2018speech}. This dataset comprises a diverse array of one-second audio clips, each featuring a spoken word. It includes a total of 105\,829 utterances encompassing 35 different words. The dataset is partitioned into 84\,843 training utterances, 9981 validation utterances, and 11\,005 test utterances. All audio files are sampled at 16 kHz and saved in WAV format with 16-bit PCM encoding.

Since our network required training and evaluation with PDM signals, we transformed the GSC dataset using a PCM to PDM conversion algorithm detailed in the following section.

\subsection{PCM to PDM}

The GSC dataset used for experiments is originally encoded using PCM. To align with the data expected from a PDM microphone, we converted it into PDM format. In the conversion process, the PCM signal undergoes initial oversampling based on the desired oversampling ratio \(\alpha\) and is then encoded using the PCM to PDM algorithm. In the following, we will first introduce the classic conversion algorithm and then present our parallelized version.

The original algorithm \cite{wiki:Pulse-density_modulation} is described in Algorithm~\ref{algo:seq} where \(x\) represents the input PCM signal, \(qe\) the quantization error and \(y\) is resulting PDM signal. In this algorithm, the input signal is normalized within the range [1, -1], with output values discretized as either 1 (indicating a pulse) or -1 (indicating no pulse).

By substituting \(y[n] \gets -1 \) with \(y[n] \gets 0\) in line 8 of the algorithm, we have modified it to accommodate input normalized between 0 and 1. The outputs are now conveniently represented as either 0 or 1, aligning with our specific context. 

The revised algorithm now essentially involves accumulating the input \(x\) into the error variable \(qe\). When the error reaches zero, a pulse is emitted, and the error is reduced by 1. This adaptation renders the algorithm equivalent to an Integrate and Fire neuron with a soft reset and a spiking threshold of 1.

Furthermore, given that the conversion algorithm is time-consuming, especially for longer input sequences (higher oversampling ratios), we have implemented a parallel over-time version of the algorithm, as outlined in Algorithm~\ref{algo:par}.
In this parallelized approach, instead of sequentially accumulating the error and resetting it upon reaching the threshold, we compute the cumulative sum of the input \(x\). A pulse is generated whenever the cumulative sum surpasses a multiple of the threshold \(th\). To identify these time points, we first perform an integral division of the signal's cumulative sum, resulting in a step curve. Subsequently, we compute the difference between consecutive samples to detect transitions in the step curve, indicating the precise moments when pulses are emitted.
This optimized version significantly accelerates the conversion process by up to 100\,000 times for the longest sequences (oversampling ratio of 64) with a Tesla V100 Graphics Processing Unit.

\begin{algorithm}
\caption{PCM to PDM Sequential Conversion}
\label{algo:seq}
\begin{algorithmic}[1]
\Function{PCM2PDM\_seq}{$\text{real}[0..s] \ x, \ \text{real} \ qe = 0$}
    \State $\text{int}[0..s] \ y$
    \For{$n$ from $0$ to $s$}
        \State $qe \gets qe + x[n]$
        \If{$qe > 0$}
            \State $y[n] \gets 1$
        \Else
            \State $y[n] \gets -1$
        \EndIf
        \State $qe \gets qe - y[n]$
    \EndFor
    \State \textbf{return} $y, qe$
\EndFunction
\end{algorithmic}
\end{algorithm}

\begin{algorithm}
\caption{PCM to PDM Parallel Conversion}
\label{algo:par}
\begin{algorithmic}[1]
\Function{PCM2PDM\_par}{$\text{real}[0..s] \ x, \ \text{int} \ th = 1$}
    \State $\text{x} \gets \text{cumsum}(\text{x})$ \Comment{Cumulative sum}
    \State $\text{x} \gets \lfloor \text{x} \div th \rfloor$  \Comment{Divide with integral result}
    \State $\text{x} \gets \text{x}[1..s] - \text{x}[0..s-1]$
    \State $\text{y} \gets \text{x} > 0$
    \State \textbf{return} $\text{y}$
\EndFunction
\end{algorithmic}
\end{algorithm}

\section{Results and Discussion}
In our evaluation, each experiment is repeated five times with random initialization and run for 150 epochs. Key metrics include classification accuracy (percentage of correctly detected words), spike rate (average number of spikes per second in the network), and the number of learned parameters.

\subsection{Ablation study}

To evaluate the effectiveness and quantify the contribution of each element incorporated in the network described in section~\ref{sec:network}, we conducted ablation studies. This involved incrementally stacking each element to assess its individual impact. The results of these experiments are presented in Table~\ref{tab:ablation_results}. All the experiments in this context are conducted with an oversampling ratio of 10.

\textit{Conv}: This represents the simplest architecture comprising only 1D convolutional layers. The baseline results achieved are 87.83\% on the training set and 77.35\% on the test set. These outcomes affirm the efficacy of 1D convolutional layers in learning diverse filters capable of extracting and processing frequency components from temporal signals, as supported by prior studies \cite{abdoli2019end}.

\textit{Conv+Rec}: This architecture incorporates recurrence to layers 3 and 4. The inclusion of recurrence yields a notable improvement, with a 9\% increase on the training set and 7\% on the test set. This enhancement is attributed to the recurrent connections' capacity to learn temporal dependencies, which was only achievable through the spiking neuron's membrane potential previously. Combining recurrence and convolution has been demonstrated to be effective \cite{arik2017convolutional}, as it leverage the strengths of both mechanisms. However, it is essential to note that the introduction of recurrence contributes to an increase in network parameters, incorporating an additional 33k parameters (comprising $2 \times 128 \times 128$ weights and $2 \times 128$ biases).

\textit{Conv+Rec+Delay}: Additionally, an axonal delay is incorporated after each of the four hidden layers. This integration results in a 1\% increase on the training set and a 2\% increase on the test set. Delays facilitate coincident spike arrival times, even when emission times differ, which can enhance network activity due to neurons' sensitivity to spike synchrony. This phenomenon has been reported to enhance network performance \cite{hammouamri2023learning}, even in scenarios where the delays are randomly fixed, as in our context.

\textit{Conv+Rec+Delay+Aug}: In this configuration, the architecture remains unchanged, but data augmentation is introduced in the training process. This augmentation involves time-shifting the audio signal with a randomly chosen value within the range [-0.3; 0.3] seconds. The outcome is a 7\% decrease in the training set but a 3\% increase in the test set. This observation underscores the enhanced generalization capacity of the network.

\begin{table}
\setlength{\tabcolsep}{3pt}
\caption{Ablation Study Results}
\label{tab:ablation_results}
\centering
\begin{tabular}{lccc}
\hline
\multirow{2}{*}{\textbf{Architecture}} & \multicolumn{3}{c}{\textbf{Accuracy (\%)}}     \\
                                       & \textbf{Train} & \textbf{Valid} & \textbf{Test} \\ \hline
\textbf{Conv}                          & 87.83±0.12 & 79.08±0.31 & 77.35±0.40             \\
\textbf{Conv+Rec}                      & 96.06±0.25 & 85.06±0.13 & 84.01±0.25           \\
\textbf{Conv+Rec+Delay}                & 97.51±0.15 & 87.39±0.20 & 86.12±0.47            \\
\textbf{Conv+Rec+Delay+Aug}            & 90.34±0.28 & 90.07±0.22 & 89.29±0.26             \\ \hline
\end{tabular}
\end{table}

\subsection{Impact of Oversampling}

\begin{figure}
\centering
\includegraphics[width=1\linewidth]{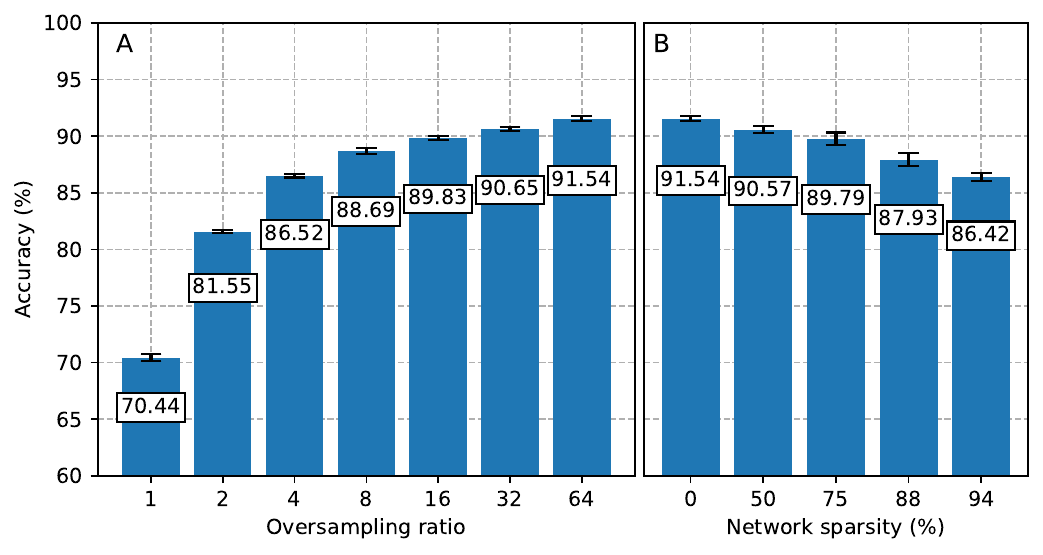}
\caption{Oversampling and network sparsity impact on classification accuracy.}
\label{img:pdm_factor_acc}
\end{figure}

In the PDM encoding scheme, the oversampling ratio is a key factor influencing the quality of the resulting encoded signal. To assess its impact on classification accuracy, we varied this parameter and observed the outcomes, as depicted in \mbox{Figure~\ref{img:pdm_factor_acc}-A}.

Initially, it is remarkable that even in the absence of oversampling, the network achieves an impressive accuracy of 70.44±0.30\%. This result underscores the system's robustness to noise, given that transitioning from a 16-bit system to a 1-bit system typically introduces increased noise by about 90 dB \cite{kite2012understanding}. Overall, the relationship between classification accuracy and the oversampling ratio follows a logarithmic pattern. Doubling the original sampling rate, for instance, results in an 11\% accuracy increase, reaching 81.55±0.13\%, while quadrupling the rate yields 86.52±0.17\%, reflecting a 5\% increase in accuracy. The best performance in this study is attained when the oversampling ratio is 64, yielding 91.54±0.23\% accuracy.

Moreover, changing the oversampling ratio has minimal impact on the network architecture, thanks to its inherent flexibility. Specifically, only the kernel size and stride of the first layer are adjusted accordingly. This leads to minor changes in network size, as indicated in Table~\ref{tab:efficiency}, and will be further discussed in the following section.

\subsection{Computational Efficiency}

To discuss the computational efficiency of the proposed approach, we focus on the following key metrics: the number of parameters, the input sampling rate (Hz or sample/s), the spike rate in the network (spike/s), and the relative spike rate (spike/sample) which is the ratio between the last two. Detailed results are presented in Table~\ref{tab:efficiency}.

As mentioned in the previous section, the oversampling ratio has a moderate impact on the number of parameters. When the ratio is 1, the network comprises 185K parameters, increasing to 210K parameters when set to 64. To further enhance network sparsity, we reduced inter-layer synaptic connectivity while maintaining a fixed oversampling ratio of 64. Instead of full connectivity ($128\times 128$), we induced sparsity levels of 50\% ($128\times 64$), 75\% ($128\times 32$), 88\% ($128\times 16$), and 94\% ($128\times 8$). This resulted in a substantial three-fold reduction in network size, down to 71K parameters. Remarkably, the resulting classification accuracy experienced only a 5\% decrease, reaching 86.42±0.13\% (\mbox{Figure~\ref{img:pdm_factor_acc}-B}).

Furthermore, the oversampling ratio directly increases the input sampling rate, reaching up to 1 MHz. Consequently, the average spiking activity in the network also increases with a growing number of input samples. For instance, at 16 kHz, the network generates 122K spikes/s, while at 1024 kHz, it produces 206K spikes/s. Notably, the spike rate increases at a slower pace than the input sampling rate, as evidenced by the ratio between the two which gives the relative spike rate. This relative spike rate is 7.67 spikes/sample for a 16 kHz input and 0.20 spikes/sample for a 1024 kHz input. In other words, for each sample of the 16 kHz input, on average, 7.67 neurons among the 512 neurons are activated (representing 1.5\%). For the 1024 kHz input, for each sample, only 0.2 neurons are activated (representing 0.04\%). This high sparsity in network activity, coupled with sparse network connectivity, holds the potential to lower energy consumption significantly when implemented on neuromorphic hardware.

\begin{table}
\setlength{\tabcolsep}{2pt}
\caption{Oversampling Ratio ($\alpha$) and Network Sparsity Impact on Number of Parameters, Input Sampling Rate (ISR), Spike Rate (SR) and Relative Spike Rate (RSR).}
\label{tab:efficiency}
\centering
\begin{tabular}{lcccc}
\hline
\multicolumn{1}{c}{\textbf{\begin{tabular}[c]{@{}c@{}} $\alpha$ (sparsity) \end{tabular}}}  &
\multicolumn{1}{c}{\textbf{\begin{tabular}[c]{@{}c@{}}\#Params \end{tabular}}}  & 
\multicolumn{1}{c}{\textbf{\begin{tabular}[c]{@{}c@{}} ISR\\(sample/s) \end{tabular}}}  & 
\multicolumn{1}{c}{\textbf{\begin{tabular}[c]{@{}c@{}}SR\\(spike/s) \end{tabular}}}  & 
\multicolumn{1}{c}{\textbf{\begin{tabular}[c]{@{}c@{}}RSR\\(spike/sample) \end{tabular}}}  \\ \hline
1 (0\%)                     & 185K           & 16K                & 122K        & 7.67                                   \\
2 (0\%)                     & 186K           & 32K                & 130K        & 4.06                                   \\
4 (0\%)                     & 187K           & 64K                & 118K        & 1.85                                   \\
8 (0\%)                     & 188K           & 128K               & 137K        & 1.07                                   \\
16 (0\%)                    & 191K           & 256K               & 158K        & 0.62                                   \\
32 (0\%)                    & 197K           & 512K               & 174K        & 0.34                                   \\
64 (0\%)                    & 210K           & 1024K              & 206K        & 0.20                                   \\ \hline
64 (50\%)                   & 136K           & 1024K              & 229K        & 0.22                                   \\
64 (75\%)                   & 99K            & 1024K              & 286K        & 0.28                                   \\
64 (88\%)                   & 81K            & 1024K              & 317K        & 0.31                                   \\
64 (94\%)                   & 71K            & 1024K              & 282K        & 0.28                                   \\ \hline
\end{tabular}
\end{table}

\subsection{Comparison with literature}

We compare our approach to results presented using the Spiking Speech Commands (SSC) dataset \cite{cramer2020heidelberg}, a spike-encoded version of GSC using an artificial model of the inner ear. We note that as the SSC dataset incorporates a time-frequency decomposition in the encoding process, less is ultimately required from the reported models to perform the KWS task.

State-Of-The-Art (SOTA) results reported for the SSC dataset in \cite{hammouamri2023learning} present network sizes ranging from 280K to 10M parameters. Interestingly, our largest network comprises 210K parameters, standing out as the smallest among the previously reported network sizes. Moreover, in terms of accuracy, the reported results for SSC vary from 50\% to 80.69\%, falling well below the 91.54±0.23\% accuracy achieved in our study, despite the more challenging end-to-end KWS task. These two metrics collectively underscore our approach as an efficient and effective strategy for neuromorphic KWS.

\section{Conclusion}

This paper introduces a novel neuromorphic Keyword Spotting approach that directly connects a Pulse Density Modulation (PDM) microphone to a Spiking Neural Network. Apart from demonstrating the correlation between PDM and the Integrate and Fire neuron, the study validates the proposed approach on the Google Speech Commands (GSC) dataset, achieving an accuracy of 91.54±0.23\%. This surpasses the state-of-the-art for the spike-encoded version of GSC dataset (SSC). Furthermore, the observed sparsity in network activity (as low as 0.04\% neuron activation) and connectivity, coupled with the elimination of preprocessing steps, emphasize the potential of the proposed approach as a promising low energy neuromorphic KWS system. 
Forthcoming studies will involve implementing this approach on a hardware device and validating the power efficiency through empirical measurements.

\section{Acknowledgements}
The authors would like to thank the \emph{Ministère de l'Économie, de l'Innovation et de l'Énergie du Québec} (MEIE) for funding their research. We would also like to thank professor \mbox{Chris Eliasmith}, Natarajan Vaidyanathan, and John Saavedra for their insights that contributed to the development of this work. This research was made possible in part by the support and computing resources provided by the Digital Research Alliance of Canada.

\bibliographystyle{IEEEtran}
\bibliography{mybib}

\end{document}